\documentclass{article} %
\usepackage[preprint]{colm2026_conference}

\usepackage{microtype}
\usepackage{hyperref}
\usepackage{url}
\usepackage{booktabs}

\usepackage{amsmath}
\usepackage{graphicx}
\usepackage{float}
\usepackage{wrapfig}
\usepackage{tcolorbox}
\usepackage{enumitem}

\usepackage{hyperref}
\usepackage{fontawesome5}

\usepackage{lineno}

\definecolor{darkblue}{rgb}{0, 0, 0.5}
\hypersetup{colorlinks=true, citecolor=darkblue, linkcolor=darkblue, urlcolor=darkblue}

\title{QuantSightBench: Evaluating LLM Quantitative Forecasting with Prediction Intervals}

\author{
  Jeremy Qin\textsuperscript{1,2,3} \qquad Maksym Andriushchenko\textsuperscript{1,2,3} \\[0.6em]
  \textsuperscript{1}ELLIS Institute Tübingen \quad
  \textsuperscript{2}Max Planck Institute for Intelligent Systems \\[0.3em]
  \textsuperscript{3}Tübingen AI Center
}

\begin{document}

\ifcolmsubmission
\linenumbers
\fi

\maketitle

\vspace{-1.5em}
\begin{center}
  \href{https://quantsightbench.com/}{%
    \fcolorbox{gray}{gray!10}{\faGlobe\ Leaderboard}%
  }
  \hspace{1em}
  \href{https://github.com/aisa-group/quantsightbench}{%
    \fcolorbox{gray}{gray!10}{\faGithub\ Code}%
  }
\end{center}

\begin{abstract}
Forecasting has become a natural benchmark for reasoning under uncertainty. Yet existing evaluations of large language models remain limited to judgmental tasks in simple formats, such as binary or multiple-choice questions. In practice, however, forecasting spans a far broader scope. Across domains such as economics, public health, and social demographics, decisions hinge on numerical estimates over continuous quantities, a capability that current benchmarks do not capture. Evaluating such estimates requires a format that makes uncertainty explicit and testable. We propose prediction intervals as a natural and rigorous interface for this purpose. They demand scale awareness, internal consistency across confidence levels, and calibration over a continuum of outcomes, making them a more suitable evaluation format than point estimates for numerical forecasting. To assess this capability, we introduce a new benchmark QuantSightBench, and evaluate frontier models under multiple settings, assessing both empirical coverage and interval sharpness. Our results show that none of the 11 evaluated frontier and open-weight models achieves the 90\% coverage target, with the top performers Gemini 3.1 Pro (79.1\%), Grok 4 (76.4\%), and GPT-5.4 (75.3\%) all falling at least 10 percentage points short. Calibration degrades sharply at extreme magnitudes, revealing systematic overconfidence across all evaluated models.
\end{abstract}

\section{Introduction} \label{sec:introduction}

Forecasting, the prediction of future outcomes under uncertainty, is foundational to decision-making across nearly every domain. In high-stakes settings such as epidemiology, climate policy, and macroeconomic planning, forecast quality directly shapes interventions that affect millions of lives. But the relevance of forecasting extends well beyond institutional contexts. Individuals routinely make forecast-dependent decisions: estimating future expenses, anticipating price changes, or gauging demand for a product. In each case, producing a good forecast requires gathering information from diverse sources, synthesizing evidence, and iteratively refining one's estimates, capabilities that are increasingly described as agentic. Large language models, with their growing ability to search, reason, and synthesize information, are already being consulted for exactly these kinds of judgments \citep{karger2025forecastbench, zeng2025futurexadvancedlivebenchmark, liu2026futurexproextendingfutureprediction}. Understanding how well they perform is therefore necessary to ensure their reliable use in decision-making, as models that convey poorly calibrated uncertainty risk silently misinforming decisions at every scale. 

Beyond its practical relevance, forecasting is also particularly well suited as a benchmark for evaluating LLM capabilities. Foresight, the ability to anticipate outcomes that have not yet occurred, is a demanding test of reasoning that integrates world knowledge, contextual understanding, and uncertainty estimation. Moreover, forecasting tasks are naturally grounded in future events, which provides two key properties for rigorous evaluation: ground truth labels are obtained automatically once outcomes are resolved, and temporal separation between training data and evaluation targets offers a principled mechanism for preventing data leakage when the benchmark is carefully constructed \citep{paleka2026pitfalls}. However, existing benchmarks evaluate forecasting through a narrow lens by primarily testing whether models can predict discrete event outcomes through binary or multiple-choice formats, which are susceptible to evaluation shortcuts \citep{chandak2025answermatchingoutperformsmultiple}, leaving out a broad class of tasks across economics, public health, and social demographics where decisions depend not on whether something will happen, but on how much.

\begin{figure}[t]
  \centering
  \includegraphics[width=\columnwidth]{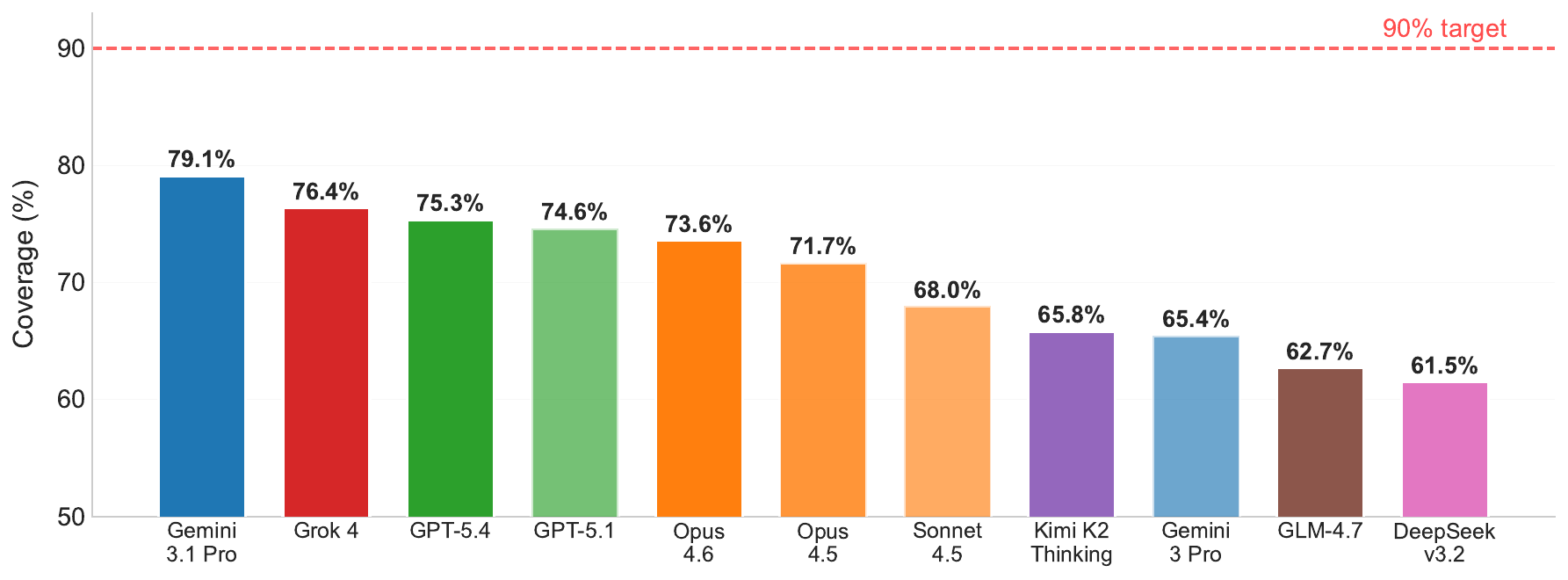}
  \caption{Empirical coverage under the agentic setting. Coverage reflects the proportion of true values captured by 90\% prediction intervals. No model achieves the target, indicating systematic overconfidence across all evaluated models.}
  \label{fig:coverage_ranking}
\end{figure}

To address this gap, we propose prediction intervals as an evaluation format for numerical forecasting in LLMs. A prediction interval is a bounded numerical range at a specified confidence level, formalizing how people naturally reason about uncertain quantities: not as point estimates, but as ranges ("between 2\% and 4\%," "around 10 to 15 million"). Notably, this intuition already surfaces in frontier models. When asked to forecast a quantity such as the population of Montreal in 2027, models like GPT-5.1 spontaneously produce interval-valued responses rather than single point estimates, suggesting that this format aligns not only with human reasoning but also with how models naturally express numerical uncertainty. Beyond being intuitive, prediction intervals offer a richer and more robust evaluation format than point forecasts by making uncertainty explicit, while remaining more practical to elicit and evaluate than full probability distributions \citep{Gneiting01032007, annurev:/content/journals/10.1146/annurev-statistics-062713-085831}. They also impose meaningful demands on model capability: producing calibrated intervals requires scale awareness, internal consistency across confidence levels, and calibration over a continuum making this task a good testbed for capability evaluation.

To this end, we make the following contributions:
\begin{itemize}
    \item We introduce QuantSightBench, a benchmark for evaluating LLM numerical forecasting capability through prediction intervals across diverse domains.
    \item We evaluate models spanning frontier and open-weight families under multiple settings (zero-shot, context-grounded, and agentic), providing a controlled comparison of how evaluation setup affects calibration.
    \item We analyze calibration across models, settings, and scales, identifying systematic overconfidence and key failure modes such as scale sensitivity.
\end{itemize}

\begin{table}[t]
    \centering
    \vspace{5pt}
    \label{tab:example-questions}
    \small
    \begin{tabular}{p{5.5cm} p{6cm} p{1cm}}
      \toprule
      \textbf{Question} & \textbf{Background} & \textbf{Answer} \\
      \midrule
      What will be the annual silver supply deficit (in million ounces) in 2025?
      & The silver market faces structural deficits when demand exceeds supply, impacting prices and industrial availability.
      & 117.6 \\
      \midrule
      What is the annual climate finance commitment agreed upon by developed nations for developing countries by 2035?
      & Developed nations have pledged financial support to help developing countries transition to clean energy and adapt to climate change.
      & 300 \\
      \midrule
      How many fatalities were confirmed in the Air India Flight AI~171 crash by October 2025?
      & Air India Flight AI~171 crashed shortly after takeoff from Ahmedabad, leading to a major aviation disaster investigation.
      & 260 \\
      \bottomrule
    \end{tabular}
    \caption{Forecasting question examples from the benchmark.}
\end{table}

\section{Related Work} \label{sec:related_works}
\subsection{Forecasting with LLMs}
Forecasting has a long history across diverse fields, from time-series methods in econometrics \citep{box2008time} and weather prediction \citep{richardson2010weather} to probabilistic forecasting in finance \citep{43f4765e-7e8a-3ad2-bff8-b0128251361a}. A growing body of work has explored the use of large language models for applications in time-series forecasting, leveraging their pattern recognition and contextual reasoning capabilities to predict structured numerical sequences \citep{jin2024timellm, gruver2023large}. More recently, judgmental forecasting, where predictions are expressed in natural language rather than derived from statistical models, has emerged as an increasingly popular benchmark for evaluating LLM reasoning under uncertainty. Platforms such as ForecastBench \citep{karger2025forecastbench}, FutureX \citep{zeng2025futurexadvancedlivebenchmark, liu2026futurexproextendingfutureprediction}, and the Metaculus AI Benchmarking Series provide dynamic leaderboards that track model performance over time. These benchmarks predominantly evaluate models on binary or multiple-choice questions sourced from prediction markets. However, this evaluation paradigm is not without limitations, \citet{paleka2026pitfalls} identified methodological concerns including data leakage and distributional biases inherent to prediction market questions.

To diversify beyond prediction markets, recent work has also explored automated question generation from news articles \citep{dai2025llmsprescientcontinuousevaluation, guan2024openepopenendedfutureevent, wang-etal-2025-openforecast}, and the OpenForecast pipeline \citep{chandak2026scalingopenendedreasoningpredict} scaled this approach through a fully automated framework that uses static news snapshots to mitigate leakage risks. Other works such as \citet{paleka2025consistencycheckslanguagemodel} also looked at different evaluation methodology such as using consistency checks that measures logical coherence across related predictions without requiring ground truth resolution. Despite this progress, evaluation across all of these efforts remains limited to purely judgmental forecasting.

\subsection{Uncertainty Estimation in LLMs}
A growing body of work has studied uncertainty quantification in LLMs, primarily in discrete prediction settings. Common approaches include token-level probabilities \citep{kadavath2022languagemodelsmostlyknow}, verbalized confidence scores \citep{xiong2024can, tian-etal-2023-just, yang2024verbalizedconfidencescoresllms}, and sampling-based methods such as semantic entropy \citep{kuhn2023semantic}. These methods have been applied to tasks such as question answering, classification, and fact verification, where quantifying model uncertainty is critical for detecting hallucinations and enabling selective prediction \citep{geng-etal-2024-survey}. A recurring finding across these studies is that LLMs tend to be overconfident, frequently assigning high confidence to incorrect predictions \citep{xiong2024can}. 

While these approaches are effective for classification and multiple-choice settings, they do not extend to expressing calibrated uncertainty over continuous quantities. Post-hoc calibration methods such as conformal prediction \citep{vovk2005algorithmic} can provide coverage guarantees over continuous outputs but wraps an existing predictor rather than evaluating whether a model can natively produce well-calibrated intervals. Prediction intervals are a well-established tool for quantifying uncertainty over continuous values in the statistical forecasting literature \citep{fc530716-e23a-3b76-8d96-2e606dcc9d52}, widely used in time-series forecasting and probabilistic regression \citep{hyndman2021forecasting}. Yet their use as an evaluation format for LLM-generated forecasts remains unexplored. Our work bridges this gap by adopting prediction intervals as the evaluation interface.

\begin{figure}[t]
  \centering
  \includegraphics[width=0.85\columnwidth]{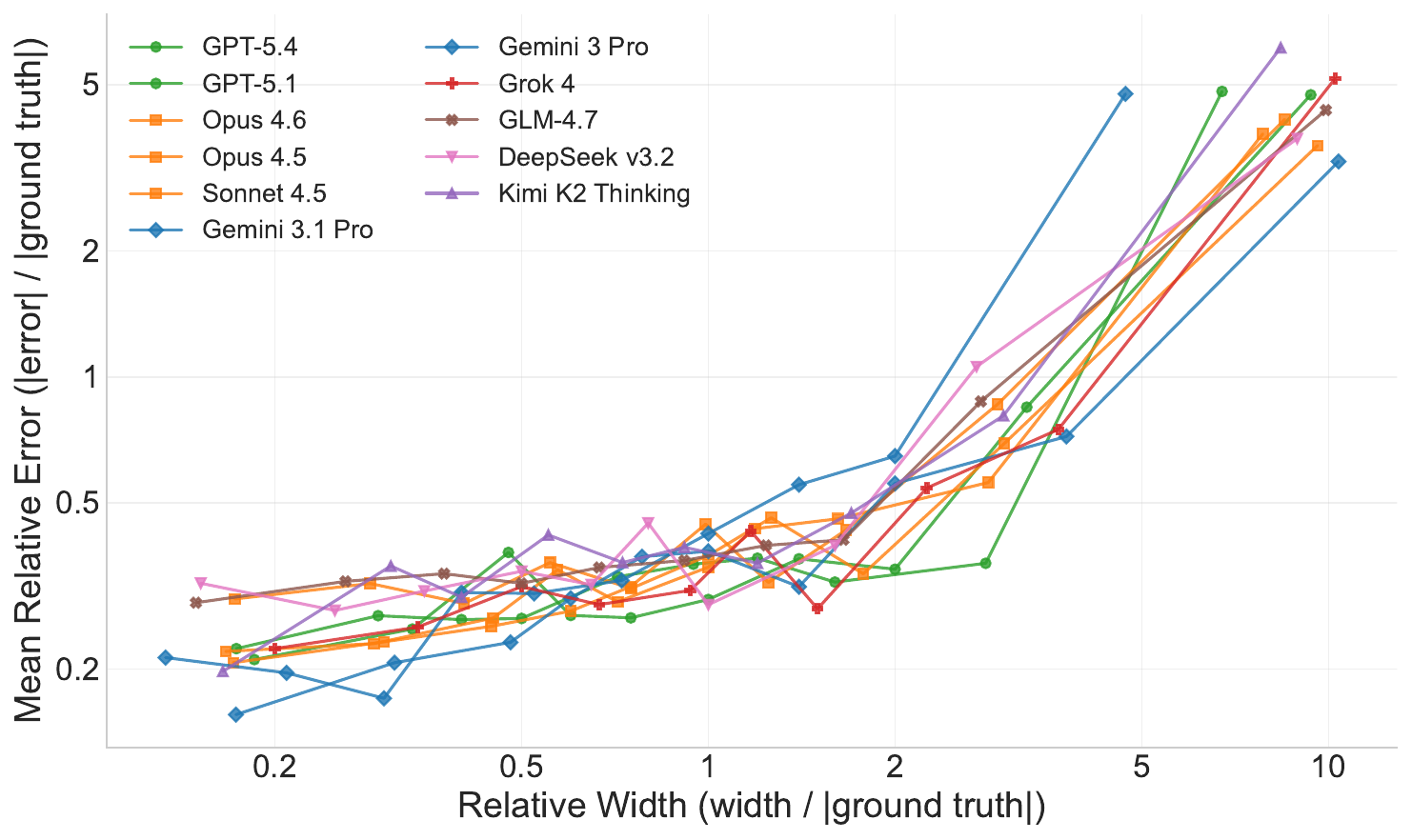}
  \caption{Relative interval width vs. mean relative error under the agentic setting. A positive trend indicates that models widen their intervals when their estimates are less accurate.}
  \label{fig:uncertainty_awareness}
\end{figure}

\section{QuantSightBench Setup} \label{sec:setup}
The following section describes the experimental setup for QuantSightBench, covering the three evaluation settings, the data construction and retrieval pipeline, and the metrics used to assess LLM numerical forecasting performance.
\subsection{Evaluation Settings}
We evaluate models under three different settings of increasing complexity. Unless otherwise specified, all evaluations are conducted at the 90\% confidence level. We analyze the impact of specifying a confidence level in the prompt in Section \ref{sec:ablation}.

\textbf{1. Zero-shot:} Models receive only the forecasting question and are asked to produce a prediction interval with no additional context. This setting tests the model's baseline forecasting reasoning from its internal knowledge. 

\textbf{2. Background-context:} Models receive the forecasting question along with relevant background information that helps in understanding important aspects of the question. This setting evaluates whether additional background information helps contextualize and improve forecasting performance.

\textbf{3. Agentic:} Models are given a retrieval tool that allows them to query and retrieve news articles from a fixed corpus. The agent can iteratively decide how many articles to retrieve and whether additional queries are needed before producing its forecast. This setting reflects a more realistic deployment scenario where the model actively gathers and synthesizes information before reasoning under uncertainty. The full agentic prompt template is shown in Figure~\ref{fig:agentic-prompt} (Appendix).

For each question, models are asked to produce a prediction interval at a specified confidence level: a lower and upper bound representing the range within which the model believes the true value will fall with the stated probability. We evaluate a range of models spanning frontier LLMs (OpenAI, Anthropic, Google) and open-weight alternatives (GLM, Kimi, DeepSeek, Grok). To ensure no leakage, we restrict our evaluation to models with a documented knowledge cutoff date prior to September 2025.

\begin{figure}[t]
  \centering
  \includegraphics[width=0.9\columnwidth]{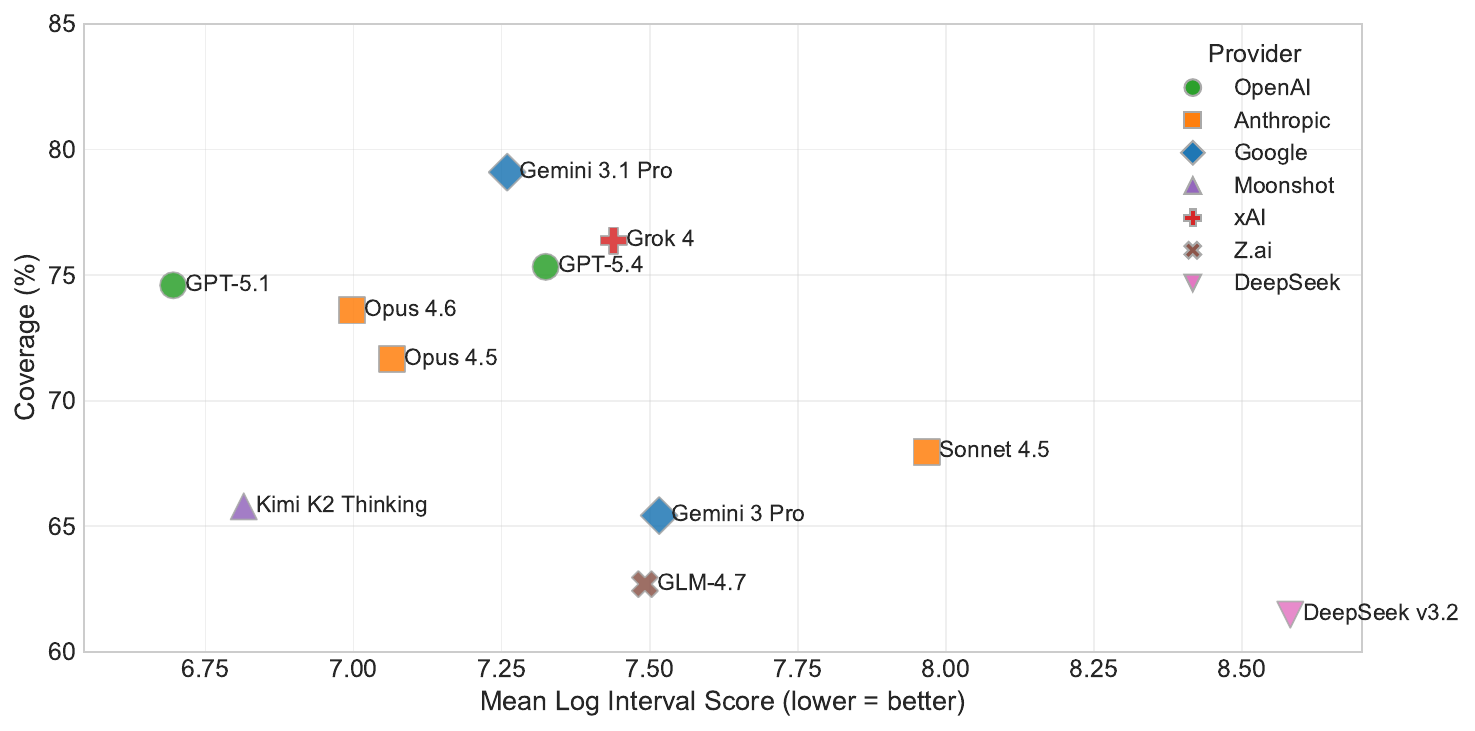}
  \caption{Coverage vs.\ MLIS under the agentic setting. Models in the upper-left corner achieve the best trade-off between calibration (high coverage) and sharpness (low MLIS).}
  \label{fig:pareto}
\end{figure}

\subsection{Data and Retrieval Pipeline}
QuantSightBench builds on the OpenForecast \citep{chandak2026scalingopenendedreasoningpredict} pipeline for news articles scraping and question generation, adapting it for numerical forecasting evaluation. We modify the question generation pipeline to retain only quantitative questions whose resolution depends on uncertain future outcomes, filtering out those answerable through straightforward calculation or factual recall from available context. To prevent data leakage and contamination, background news articles are drawn from January to August 2025, while forecasting questions are generated from articles with resolution dates between September 2025 and January 2026.

The retrieval database comprises approximately 320,000 news articles from diverse international sources including Forbes, CNN, Hindustan Times, Deutsche Welle, and Irish Times. The corpus is de-duplicated and chunked into 512-token segments, each embedded with OpenAI's text-embedding-3-large. At evaluation time, the most relevant chunks are retrieved and provided to the model as context. The final benchmark consists of 1,000 forecasting questions spanning diverse domains. Examples of the data are shown in Table \ref{tab:example-questions} and the domains covered are presented in Table \ref{tab:domain-distribution}. We refer to \citet{chandak2026scalingopenendedreasoningpredict} for full details on the question generation, ground truth extraction, and validation procedures for the data used in our experiments.

\subsection{Evaluation Metrics}

We mainly assess prediction interval quality along two dimensions:

\textbf{Coverage.} The proportion of questions for which the ground truth falls within the predicted interval. A well-calibrated model producing $1 - \alpha$ confidence intervals should achieve empirical coverage close to $1 - \alpha$. Thus, for a set of \textit{N} questions, coverage at confidence level $1 - \alpha$ is defined as:
\begin{equation}
    \text{Coverage} = \frac{1}{N} \sum_{i=1}^{N} \mathbb{1}(l_i \leq y_i \leq u_i)
\end{equation}

\textbf{Mean Log Interval Score (MLIS).} While coverage measures calibration, it does not penalize uninformative intervals: a model that always predicts $(-\infty, +\infty)$ achieves perfect coverage. To jointly assess calibration and sharpness, we use the Winkler interval score \citep{fc530716-e23a-3b76-8d96-2e606dcc9d52}, a proper scoring rule which penalizes both interval width and miscoverage. A scoring rule is said to be \emph{proper} if it is minimized in expectation when the forecaster reports their true beliefs, ensuring that the optimal strategy is honest uncertainty estimation rather than strategic hedging. For a single prediction interval at confidence level $1 - \alpha$, the interval score is defined as:

\begin{equation}
    S_\alpha(l_i, u_i, y_i) = (u_i - l_i) + \frac{2}{\alpha}(l_i - y_i)\mathbb{1}(y_i < l_i) + \frac{2}{\alpha}(y_i - u_i)\mathbb{1}(y_i > u_i)
\end{equation}

where the first term penalizes interval width and the remaining terms impose a penalty proportional to the distance by which the ground truth falls outside the interval. However, numerical forecasting span widely different scales. Applying the interval score directly would allow large-scale questions to dominate the aggregate metric. To address this, we apply the interval score to log-transformed values, yielding the \textbf{Mean Log Interval Score}:

\begin{equation}
    \text{MLIS} = \frac{1}{N} \sum_{i=1}^{N} \left[ (\log u_i - \log l_i) + \frac{2}{\alpha}(\log l_i - \log y_i)\mathbb{1}(y_i < l_i) + \frac{2}{\alpha}(\log y_i - \log u_i)\mathbb{1}(y_i > u_i) \right]
\end{equation}

The log transformation serves two purposes. First, it normalizes across scales, ensuring that no single domain or magnitude range disproportionately influences the aggregate score. Second, because the logarithm is strictly monotonic and applied consistently to both predictions and ground truth, the resulting score inherits the properness of the original Winkler score, preserving the incentive to report honest uncertainty estimates.

\section{Results} \label{sec:results}

We present the main results under the agentic setting, which reflects the most realistic scenario where models actively retrieve and synthesize information for their forecasts. We analyze coverage and sharpness across models, examine how ground truth scale and number of retrieval iterations affect performance, and investigate whether models modulate their interval width in response to question difficulty.

\begin{figure}[t]
  \centering
  \includegraphics[width=\columnwidth]{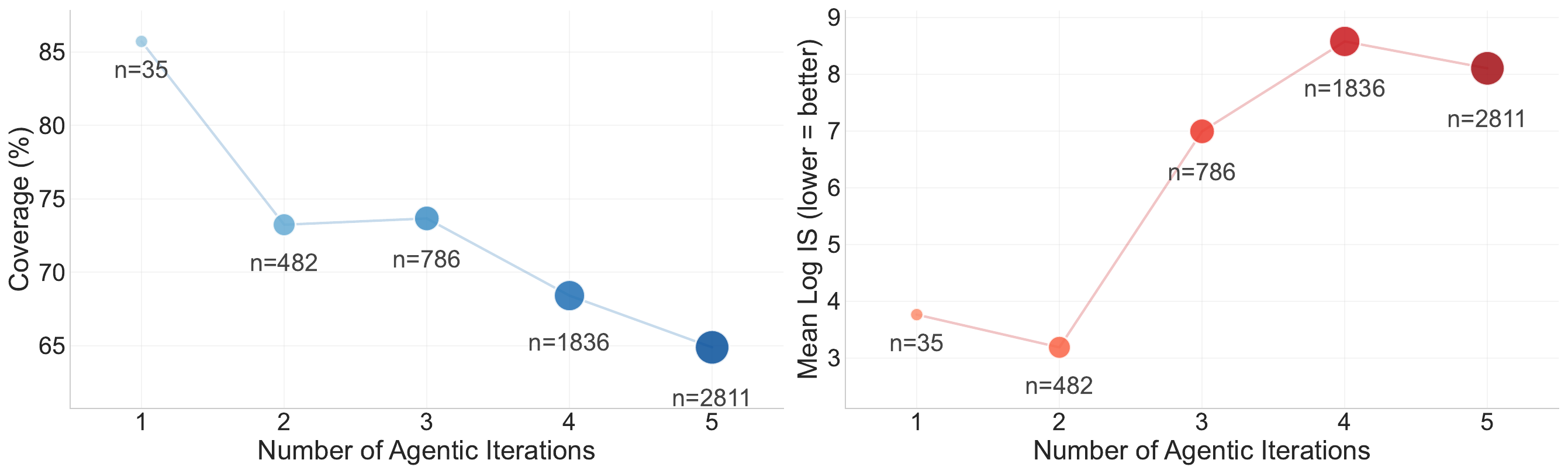}
  \caption{Coverage and MLIS by agentic retrieval iterations across all models. Questions requiring more iterations tend to be harder, yielding lower coverage and higher MLIS.}
  \label{fig:agentic_iterations}
\end{figure}

\subsection{Prediction Interval Coverage}
Figure \ref{fig:coverage_ranking} reports empirical coverage at the 90\% confidence level across all models. No model achieves the nominal 90\% target. Gemini 3.1 Pro leads with 79.1\% coverage, followed by Grok 4 (76.4\%) and GPT-5.4 (75.3\%). At the lower end, DeepSeek v3.2 (61.5\%) and GLM-4.7 (62.7\%) fall nearly 30 percentage points short of the target. This consistent shortfall across all models indicates systematic overconfidence: models produce intervals that are too narrow relative to their stated confidence level.

Coverage results are broadly consistent with model updates within the same family, with newer generations outperforming their predecessors across OpenAI, Anthropic, and Google model lines. This suggests that improvements in general model capability translate, at least partially, to better calibrated numerical forecasting. We also observe a noticeable gap between frontier proprietary models and open-weight models. This gap suggests that numerical forecasting calibration may benefit from capabilities or training procedures that are not yet fully reflected in current open-weight offerings. 

\subsection{Prediction Interval Quality}
Coverage alone does not fully characterize forecast quality, as trivially wide intervals can achieve high coverage without providing actionable information. Figure \ref{fig:pareto} plots coverage against MLIS, jointly capturing calibration and sharpness. Well-calibrated and strong models should achieve high coverage with low MLIS, indicating intervals that are both reliable and informatively tight. Gemini 3.1 Pro and GPT-5.1 come closest to this profile. GPT-5.1 is particularly notable: despite slightly lower coverage (74.6\%) than GPT-5.4 (75.3\%), it achieves a meaningfully lower MLIS. This indicates that GPT-5.1 is better at concentrating its intervals around plausible outcomes rather than compensating for uncertainty with wider ranges. The ranking along MLIS does not always mirror the coverage ranking. Sonnet 4.5, for instance, achieves moderate coverage (68.0\%) but with a relatively high MLIS, indicating that its intervals are wide yet still fail to cover a substantial fraction of outcomes. 

\begin{table}[t]
      \centering
      \vspace{5pt}
      \label{tab:reasoning-effort}
      \small
      \begin{tabular}{l cc cc cc cc}
        \toprule
        & \multicolumn{2}{c}{\textbf{GPT-5.1}} & \multicolumn{2}{c}{\textbf{Opus 4.5}} & \multicolumn{2}{c}{\textbf{Sonnet 4.5}} & \multicolumn{2}{c}{\textbf{Gemini 3 Pro}} \\
        \cmidrule(lr){2-3} \cmidrule(lr){4-5} \cmidrule(lr){6-7} \cmidrule(lr){8-9}
        \textbf{Effort} & Cov. & MLIS & Cov. & MLIS & Cov. & MLIS & Cov. & MLIS \\
        \midrule
        Low    & 77.10 & 7.43 & 65.36 & 7.45 & 69.93 & 8.35 & 63.10 & 8.62 \\
        Medium & \textbf{78.38} & \textbf{6.97} & 69.72 & 7.05 & 71.35 & 8.12 & -- & -- \\
        High   & 75.98 & 7.42 & \textbf{72.55} & \textbf{6.74} & \textbf{72.66} & \textbf{7.80} & \textbf{65.03} & \textbf{8.29} \\
        \bottomrule
      \end{tabular}
      \caption{Effect of reasoning effort on coverage and MLIS. Higher effort generally improves both metrics, except for GPT-5.1 which peaks at medium. Bold denotes best setting per model; dashes indicate unsupported configurations.}
    \end{table}

These observations highlight that miscalibration manifests in qualitatively different ways across models. Some models produce narrow intervals that fail to achieve adequate coverage, reflecting overconfidence, while others produce wide intervals that still miss the target. This distinction is important for understanding where current models fail and what improvements are needed for reliable numerical forecasting.

\subsection{Effect of Scale}
Figures \ref{fig:coverage_by_scale} and \ref{fig:mlis_by_scale} present coverage and MLIS broken down by the magnitude of the ground truth value. Coverage degrades consistently as magnitude increases: in the 1–10 range, most models achieve coverage above 80\%, while by the 100K+ range most fall below 65\% and several below 50\%. The rate of decline varies across models. The MLIS heatmap corroborates this trend. Interval scores are lowest in the 1–10 and 100–1K ranges and deteriorate at both extremes. At the low end, the 0–1 range shows elevated MLIS for several models, likely because fractional quantities such as rates and percentages require reasoning about proportions rather than absolute magnitudes. At the high end, MLIS increases dramatically, exceeding 10 for most models and reaching 23.7 for Gemini 3 Pro, indicating that models struggle to produce appropriately scaled intervals for very large quantities. These findings suggest that scale awareness is a key bottleneck in forecasting.

\subsection{Effect of Agentic Iterations}
Figure \ref{fig:agentic_iterations} examines how the number of agentic iterations relates to performance, aggregated across all models. Coverage decreases monotonically with the number of iterations, from approximately 86\% at one iteration to 65\% at five, while MLIS increases sharply over the same range. This pattern may seem counterintuitive, as more retrieval and reasoning might be expected to improve forecasts. However, it likely reflects question difficulty rather than a detrimental effect of iteration. The agent's decision to iterate further is itself a signal that the question is harder to resolve, involving quantities that are more uncertain or less directly retrievable from available sources. Conversely, easier questions tend to be resolved in fewer iterations, naturally yielding better calibration and sharper intervals. 

\subsection{Interval Width Adaptation}

Figure \ref{fig:uncertainty_awareness} plots relative interval width (width divided by ground truth magnitude) against mean relative error, where the error is computed using the interval midpoint as the point estimate. The normalization by ground truth magnitude controls for scale effects, ensuring that the observed trends reflect genuine modulation of uncertainty rather than artifacts of question magnitude. A well-calibrated model should exhibit a positive relationship: when its estimate is further from the true value, its intervals should be correspondingly wider to maintain coverage. All models exhibit this positive trend, suggesting that models do possess some degree of uncertainty modulation, adjusting their interval width in response to question difficulty. However, the relationship varies considerably across models, indicating that this awareness is far from well-calibrated.

\section{Ablation Studies} \label{sec:ablation}
We now examine how different experimental choices affect calibration and interval quality, including prompt setting, reasoning effort, and target confidence level.

\subsection{Effect of Prompt Setting}
Figure \ref{fig:prompt_setting} compares coverage and MLIS across the three different settings. Providing background context generally improves both coverage and interval quality compared to the zero-shot setting, indicating that grounding information helps models produce better calibrated forecasts. The agentic setting further improves performance, with the most pronounced gains observed for open-weight models such as GLM-4.7 and DeepSeek v3.2. Frontier models, by contrast, show smaller improvements from agentic retrieval, as they already perform relatively well under simpler prompting settings. This asymmetry suggests that the primary limitation of weaker models is information access: when given the ability to retrieve relevant context, they close part of the gap with frontier models. For stronger models, the diminishing returns from additional retrieval suggest that much of the useful signal may already be captured from the provided context, and that improving calibration may require more than simple access to recent information, such as structured reasoning over historical baselines or domain-specific priors.

\begin{figure}[t]
  \centering
  \includegraphics[width=\columnwidth]{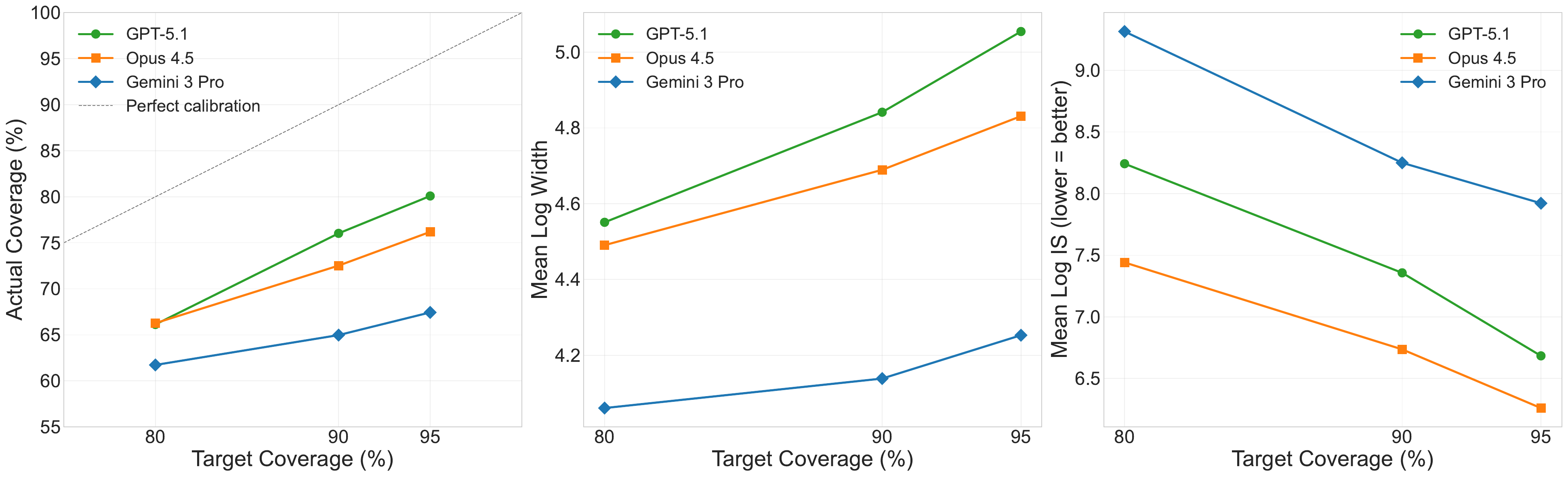}
  \caption{Calibration, interval width, and MLIS across target confidence levels (80\%, 90\%, 95\%) under the background-context setting. All models are undercoverage relative to the diagonal, but MLIS improves at higher targets as intervals widen appropriately.}
  \label{fig:confidence_levels}
\end{figure}

\subsection{Effect of Reasoning Effort}
Table \ref{tab:reasoning-effort} reports coverage and MLIS across three levels of reasoning effort (low, medium, high) under the background-context setting. Increased reasoning effort generally improves both calibration and sharpness. Opus 4.5 benefits the most, with coverage increasing from approximately 65\% to 73\% and MLIS decreasing consistently, indicating that extended reasoning helps produce both better calibrated and sharper intervals. Sonnet 4.5 shows a similar but more modest improvement. In contrast, GPT-5.1 shows minimal gains, suggesting that for already performant models, additional reasoning tends to sharpen intervals without necessarily improving coverage, as increased confidence is not always warranted. While the magnitude of improvement varies across models, these results indicate that reasoning effort is a useful lever for improving calibration, particularly for models that have not yet saturated their forecasting capability.

\subsection{Effect of Confidence Levels}
Figure \ref{fig:confidence_levels} evaluates calibration across three target confidence levels (80\%, 90\%, 95\%) under the background-context setting for GPT-5.1, Opus 4.5, and Gemini 3 Pro. All three models fall below perfect calibration at every level, with the gap widening at higher confidence targets, indicating that models struggle disproportionately to produce appropriately wide intervals when more coverage is requested. Models do widen their intervals in response to higher confidence levels, but insufficiently to achieve the requested coverage. Interestingly, MLIS improves at higher confidence levels for Opus 4.5 and GPT-5.1, as the wider intervals reduce miscoverage penalties enough to offset the increased width penalty. Gemini 3 Pro shows the opposite trend, suggesting a less efficient interval widening strategy.

We further examine the effect of specifying a confidence level in the prompt (Table \ref{tab:no-conf-ablation}). When no confidence level is specified, models must determine interval width without explicit guidance. Both coverage and MLIS degrade substantially: GPT-5.4 drops from 75.33\% to 68.24\% coverage with MLIS rising from 7.32 to 11.44, and Opus 4.6 shows a similar pattern, falling from 73.60\% to 66.99\% coverage. Without a specified confidence level, models tend to produce narrower and less calibrated intervals, indicating that confidence level specification serves as an important conditioning signal for interval quality.

\begin{table}[t]
      \centering
      \vspace{5pt}
      \label{tab:no-conf-ablation}
      \small
      \begin{tabular}{l cc cc}
        \toprule
        & \multicolumn{2}{c}{\textbf{GPT-5.4}} & \multicolumn{2}{c}{\textbf{Opus 4.6}} \\
        \cmidrule(lr){2-3} \cmidrule(lr){4-5}
        \textbf{Confidence Instruction} & Coverage & MLIS & Coverage & MLIS \\
        \midrule
        None (inherent) & 68.24 & 11.44 & 66.99 & 10.71 \\
        90\% specified   & \textbf{75.33} & \textbf{7.32} & \textbf{73.60} & \textbf{7.00} \\
        \bottomrule
      \end{tabular}
      \caption{Effect of specifying a 90\% target confidence level versus leaving it unspecified (inherent) in the agentic prompt. Explicit confidence instructions improve both coverage and MLIS for both models. Bold denotes the better setting.}
    \end{table}

\section{Limitations} \label{sec:limitations}
While QuantSightBench provides a controlled evaluation environment, several aspects could be extended. The agentic setting relies on a fixed retrieval corpus, which prevents temporal leakage but does not capture the dynamic nature of real-world forecasting where information evolves over time and forecasters iteratively update their estimates as new evidence becomes available. Extending the benchmark to a live setting with web search access and support for forecast updates would more closely simulate practical forecasting conditions. An interesting direction for future work is leveraging QuantSightBench for training better forecasting agents. The Mean Log Interval Score, as a proper scoring rule, is naturally suited as a reward signal for reinforcement learning, enabling models to be post-trained specifically for calibrated numerical forecasting.

\section{Conclusion} \label{sec:conclusion}
We introduced QuantSightBench, a benchmark for evaluating LLM numerical forecasting through prediction intervals. By shifting evaluation from discrete event prediction to continuous quantities, QuantSightBench addresses a gap in existing forecasting benchmarks and captures a more decision-relevant form of reasoning under uncertainty. Our evaluation of frontier models across multiple settings reveals that all models are systematically miscalibrated at the 90\% confidence level, with even the best model falling over 10 percentage points short of nominal coverage. Calibration generally improves across model generations, and agentic retrieval benefits weaker models more than stronger ones. However, performance degrades substantially for quantities at extreme magnitudes. These results suggest that while LLMs exhibit meaningful uncertainty modulation, adjusting interval widths in response to problem difficulty, significant improvements in calibration are needed before they can be reliably deployed for quantitative decision support.

\section*{Acknowledgements}
MA acknowledges financial support from Coefficient Giving.

\section*{LLM Usage}
We acknowledge the use of coding agents to assist in building the evaluation framework. All generated code was reviewed and validated by the authors. LLMs were also used to refine the writing of this paper for better clarity and coherence only.

\bibliography{colm2026_conference}
\bibliographystyle{colm2026_conference}

\pagebreak

\appendix
\section{Appendix}
\subsection{Forecasting Questions Domains}
Table~\ref{tab:domain-distribution} summarizes the distribution of the 1000 forecasting questions in our evaluation set. To characterize domain coverage, we assigned each question to one of eight broad categories using gpt-4.1-mini as a classifier. The distribution reflects the composition of forecasting content found on the open web via CommonCrawl
\begin{table}[H]
      \centering
      \vspace{5pt}
      \label{tab:domain-distribution}
      \small
      \begin{tabular}{l r r}
        \toprule
        \textbf{Domain} & \textbf{Count} & \textbf{\%} \\
        \midrule
        Business, Finance \& Technology   & 258 & 25.8 \\
        Politics \& Geopolitics          & 185 & 18.5 \\
        Infrastructure \& Transport      & 122 & 12.2 \\
        Culture \& Entertainment         & 113 & 11.3 \\
        Crime \& Legal                   & 101 & 10.1 \\
        Sports                           &  96 &  9.6 \\
        Science, Health \& Environment   &  70 &  7.0 \\
        Education                        &  55 &  5.5 \\
        \midrule
        \textbf{Total}                   & \textbf{1000} & \textbf{100.0} \\
        \bottomrule
      \end{tabular}
      \caption{Distribution of the 1,000 sampled forecasting questions across 8 domains. Questions were extracted from CommonCrawl using the OpenForesight pipeline.}
    \end{table}
    
\subsection{Model Performance across Settings}
Below, we show coverage and MLIS for all models across the zero-shot, context, and agentic prompt settings. The agentic setting yields the largest gains for open-weight models, while frontier models show diminishing returns from additional retrieval. A detailed discussion of these results is provided in Section \ref{sec:ablation}.
\begin{figure}[H]
  \centering
  \includegraphics[width=0.85\columnwidth]{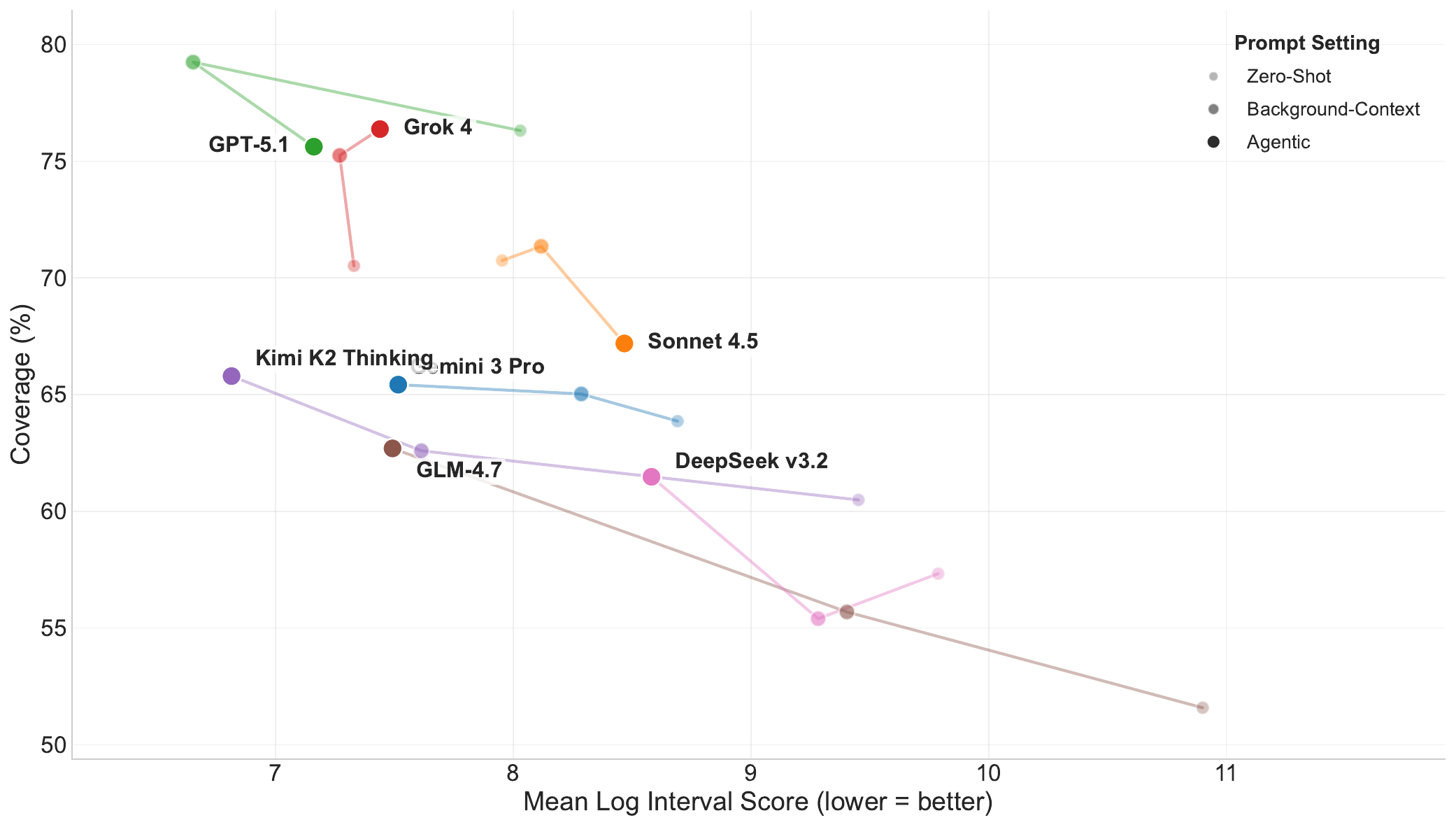}
  \caption{Coverage and MLIS across the zero-shot, context, and agentic prompt settings for all models. Models in the upper-left corner achieve the best trade-off between calibration (high coverage) and sharpness (low MLIS)}
  \label{fig:prompt_setting}
\end{figure}

\subsection{Model Performance across Ground Truth Scale}
Here, we break down coverage and MLIS by the magnitude of the ground truth value. Coverage degrades steadily with increasing magnitude, while MLIS is elevated at both extremes, suggesting that scale awareness is a key bottleneck in forecasting. More discussion is provided in Section \ref{sec:results}.
\begin{figure}[H]
  \centering
  \includegraphics[width=0.85\columnwidth]{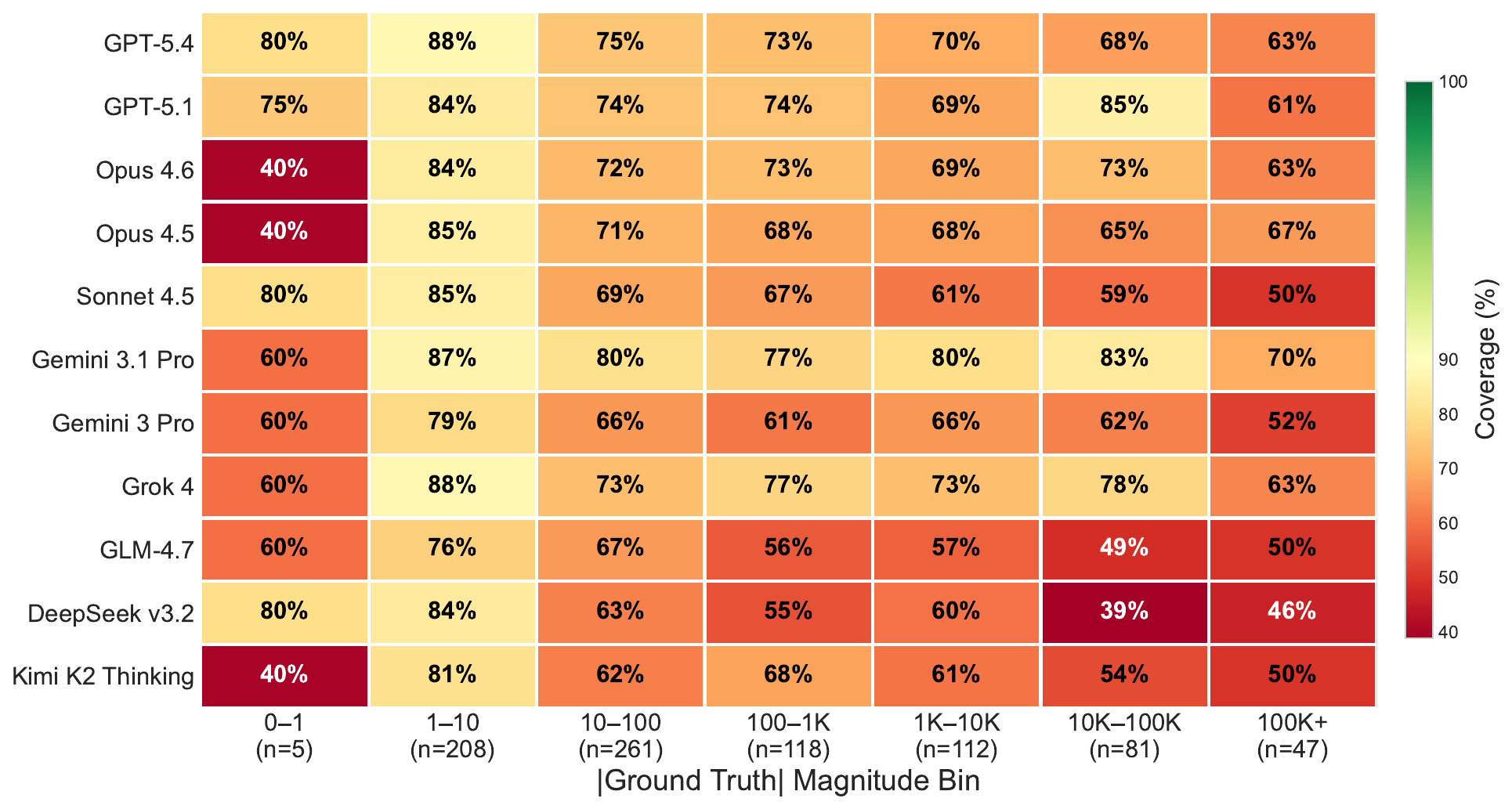}
  \caption{Coverage broken down by the magnitude of the ground truth value for all models.}
  \label{fig:coverage_by_scale}
\end{figure}

\begin{figure}[H]
  \centering
  \includegraphics[width=0.9\columnwidth]{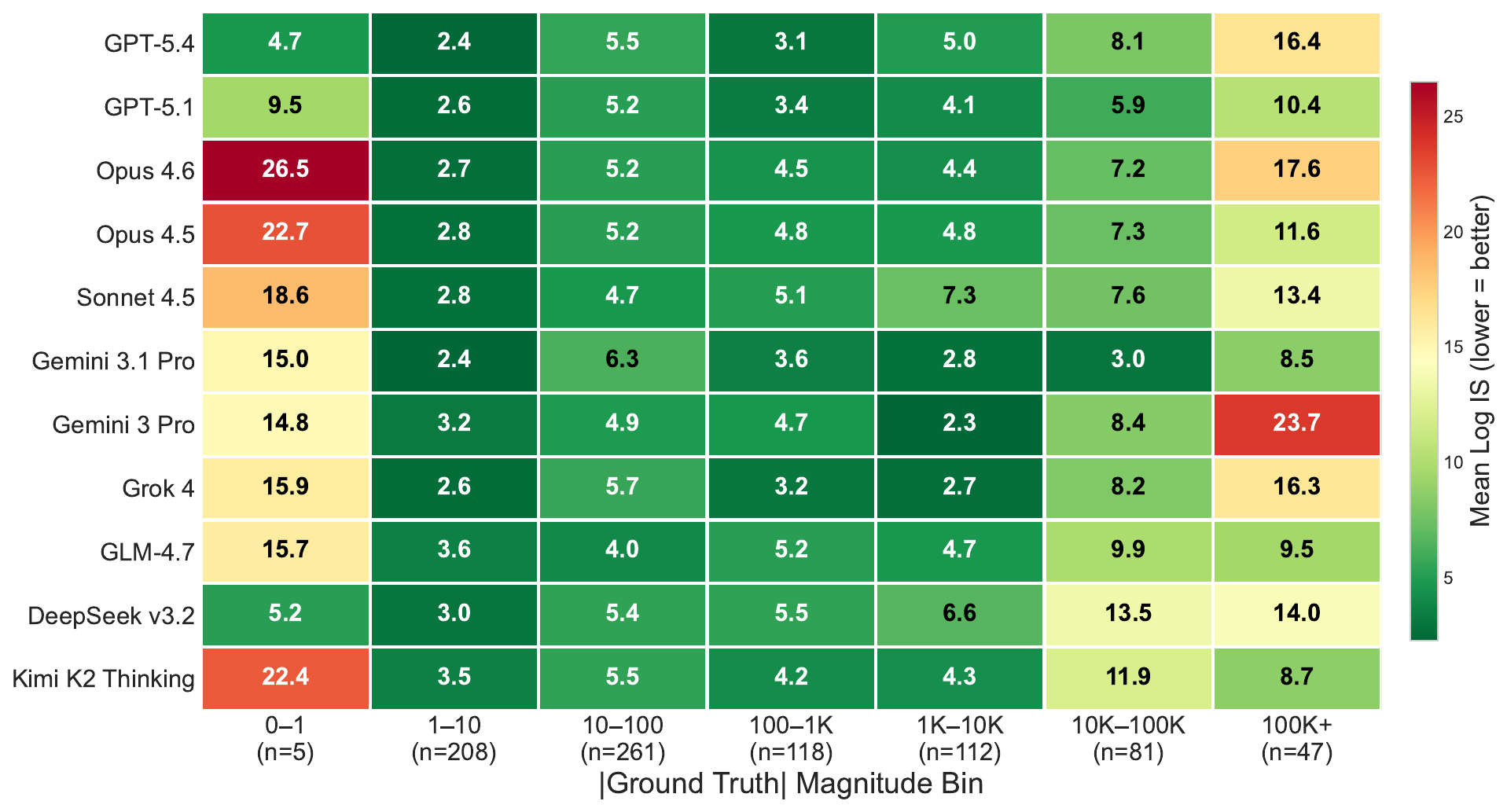}
  \caption{MLIS broken down by the magnitude of the ground truth value for all models.}
  \label{fig:mlis_by_scale}
\end{figure}

\subsection{Concrete Example of Prediction Interval in Frontier Models}
The following example illustrates GPT-5.1 response when prompted to forecast the population of Montreal in 2027. The model not only produces a point estimate but also spontaneously generates a prediction interval, suggesting that this capability is already present in frontier models without explicit instruction.
\begin{figure}[H]
  \centering
  \includegraphics[width=0.9\columnwidth]{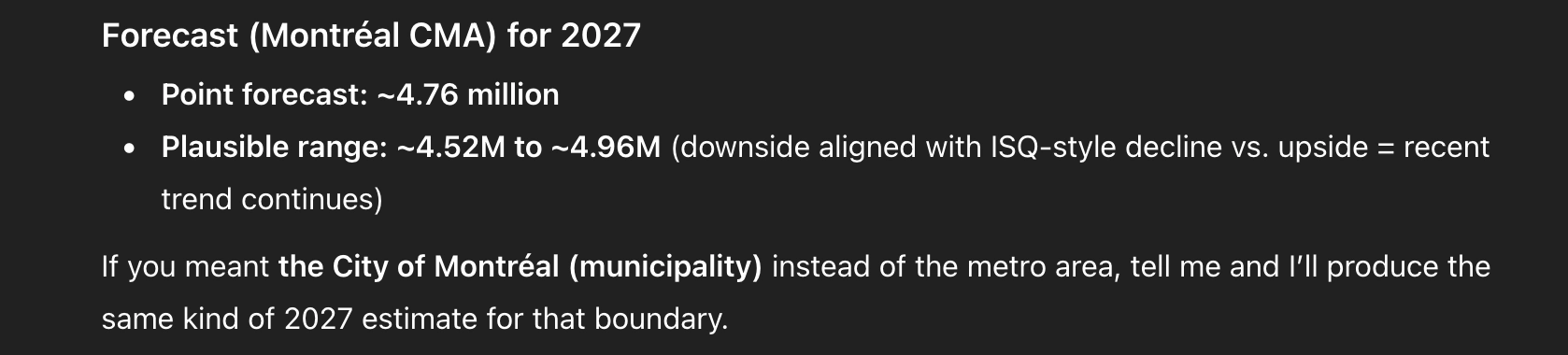}
  \caption{Forecasting example with GPT-5.1}
  \label{fig:gpt_example}
\end{figure}

\subsection{Prompt Details}
Below, we show the full prompt template used in the agentic setting.

\begin{figure}[H]
  \centering
  \begin{tcolorbox}[
    colback=blue!3, colframe=blue!40!black,
    fonttitle=\bfseries\small,
    title=Agentic Forecasting Prompt,
    boxrule=0.5pt, arc=2pt,
    left=4pt, right=4pt, top=4pt, bottom=4pt,
    fontupper=\small
  ]
  \textbf{System:} \textit{You are a calibrated forecasting assistant with access to a news article search tool. Decide whether to search for relevant articles to inform your prediction. Match the units specified
  in the resolution criteria.}
  \vspace{4pt}
  \hrule
  \vspace{4pt}
  \textbf{User:}
  You are a forecasting expert tasked with predicting a future outcome. You have access to a search tool that can retrieve relevant news articles to help inform your prediction. \\
  \\
  \textbf{Question Title:} \texttt{\{question\_title\}}\\
  \textbf{Question Background:} \texttt{\{background\}}\\
  \textbf{Resolution Criteria:} \texttt{\{resolution\_criteria\}} \\
  \\
  \textbf{AVAILABLE TOOL:}\\
  You have access to the \texttt{search\_articles} tool which searches a database of news articles.
  \begin{itemize}[nosep, leftmargin=12pt]
    \item You can search multiple times with different queries
    \item Each search returns 1--5 articles (specify \texttt{num\_results})
    \item Only articles published before the resolution date are returned
  \end{itemize}
  \vspace{7pt}
  \textbf{STRATEGY:}
  \begin{enumerate}[nosep, leftmargin=12pt]
    \item Assess what information you need. Think about what data points would reduce uncertainty.
    \item If you need external data, use the \texttt{search\_articles} tool.
    \item Evaluate relevance for \textbf{each} retrieved article: mark as \textsc{relevant} or \textsc{not relevant} with justification.
    \item Decide whether additional searches are needed.
    \item Base prediction \textbf{only} on relevant articles; if none found, rely on prior knowledge with wide intervals.
  \end{enumerate}
  \vspace{7pt}
  \textbf{PREDICTION REQUIREMENTS:}\\
  Provide a \texttt{\{probability\_level\}} prediction interval (lower, median, upper) in the same units as the resolution criteria. \\
  \\
  \textbf{OUTPUT FORMAT:}\\[2pt]
  \texttt{<lower>}\textit{NUMBER}\texttt{</lower>}\\
  \texttt{<median>}\textit{NUMBER}\texttt{</median>}\\
  \texttt{<upper>}\textit{NUMBER}\texttt{</upper>}
  \end{tcolorbox}
  \caption{Agentic forecasting prompt template. The model receives a question with background context and resolution criteria, and can iteratively call a \texttt{search\_articles} tool to retrieve relevant news
  articles before producing a prediction interval.}
  \label{fig:agentic-prompt}
\end{figure}

\end{document}